\documentclass{article}
\usepackage[utf8]{inputenc}

\usepackage{pgfplots}
\pgfplotsset{compat=1.10}
\usepgfplotslibrary{fillbetween}

\usepackage{algorithm}
\usepackage{algorithmic}
\usepackage{amssymb}
\usepackage{comment}
\usepackage{graphicx}
\usepackage{amsthm}
\usepackage{makecell}

\usepackage{url}

\author{
  Suwanviwatana Kananat \\
  \texttt{s.kananat@jaist.ac.jp}
  \and
  Hiroyuki Iida \\
  \texttt{iida@jaist.ac.jp}
}

\title{
First Results from Using Game Refinement Measure and Learning Coefficient in \textsc{Scrabble}\thanks{
Scrabble\textsuperscript{\textregistered}
is a registered trademark. All intellectual property rights in and to the game are owned in the USA by Hasbro Inc., in Canada by Hasbro Canada Corporation, and throughout the rest of the world by J.W. Spear \& Sons Limited of Maidenhead, Berkshire, England, a subsidiary of Mattel Inc.}
}

\begin{document}

\maketitle

\begin{abstract}
This paper explores the entertainment experience and learning experience in \textsc{Scrabble}. It proposes a new measure from the educational point of view, which we call {\it learning coefficient}, based on the balance between the learner's skill and the challenge in \textsc{Scrabble}.  
\textsc{Scrabble} variants, generated using different size of board and dictionary, are analyzed with two measures of game refinement and learning coefficient. The results show that 13x13 \textsc{Scrabble} yields the best entertainment experience and 15x15 (standard) \textsc{Scrabble} with 4\% of original dictionary size yields the most effective environment for language learners. Moreover, 15x15 \textsc{Scrabble} with 10\% of original dictionary size has a good balance between entertainment and learning experience.
\end{abstract}


\section{Introduction}
\label{section:introduction}
\textsc{Scrabble} has been played for a long while in various settings, e.g. as a friendly game among friends or household members, in competitive matches and also as a language learning tool. Using game refinement theory \cite{paper:mathematical_theory_of_game_refinement}, we have discovered that \textsc{Scrabble} has fun game aspect over educational aspect \cite{paper:gamification_and_scrabble}. This paper is an attempt to enhance the \textsc{Scrabble} with learning aspect.
Emotional excitement or mental engagement in games is the subject of game refinement theory. Early work in this direction has been carried out by Iida et al. \cite{paper:application_of_game_refinement_to_mah_jong}, while constructing a logistic model based on game outcome uncertainty to measure the attractiveness and sophistication of games, known as game refinement theory. Although many efforts have been devoted to the study of scoring sports and boardgames, \textsc{Scrabble} also has an educational aspect which requires extra dimension to explore.

The structure of the paper is as follows.
Section~\ref{section:scrabble} presents the basic rules of \textsc{Scrabble}.
Section~\ref{section:game_refinement_measure} and Section~\ref{section:education_dimension} describe \textsc{Scrabble} in two distinct dimensions of measurement from the perspective of entertainment and education, respectively. 
Section~\ref{section:assessment_and_discussion} presents the assessment using the swing model, thus discusses the results of the analysis, 
and concluding remarks are given in Section~\ref{section:conclusion}. 
\section{Scrabble}
\label{section:scrabble}
\textsc{Scrabble} is a word anagram game in which 2 to 4 players competitively score points by placing tiles, each bearing a single letter, onto a 15x15 board. The standard board is shown in Figure~\ref{table:scrabble_board}. The tiles must form words that are accepted by the dictionary, in either the vertical or horizontal direction in a crossword style.
\begin{figure}[ht]
    \centering
    \begin{tabular}{|p{0.4cm}|p{0.4cm}|p{0.4cm}|p{0.4cm}|p{0.4cm}|p{0.4cm}|p{0.4cm}|p{0.4cm}|p{0.4cm}|p{0.4cm}|p{0.4cm}|p{0.4cm}|p{0.4cm}|p{0.4cm}|p{0.4cm}|} 
    \hline
    $3W$ & & &$2L$ & & & &$3W$ & & & &$2L$ & & &$3W$ \\
    \hline
    &$2W$ & & & &$3L$ & & & &$3L$ & & & &$2W$ & \\
    \hline
    & &$2W$ & & & &$2L$ & &$2L$ & & & &$2W$ & & \\
    \hline
    & & &$2W$ & & & &$2L$ & & & &$2W$ & & & \\
    \hline
    & & & &$2W$ & & & & & &$2W$ & & & & \\
    \hline
    &$3L$ & & & &$3L$ & & & &$3L$ & & & &$3L$ & \\
    \hline
    & &$2L$ & & & &$2L$ & &$2L$ & & & &$2L$ & & \\
    \hline
    $3W$ & & &$2L$ & & & &$2W$ & & & &$2L$ & & &$3W$ \\
    \hline
    & &$2L$ & & & &$2L$ & &$2L$ & & & &$2L$ & & \\
    \hline
    &$3L$ & & & &$3L$ & & & &$3L$ & & & &$3L$ & \\
    \hline
    & & & &$2W$ & & & & & &$2W$ & & & & \\
    \hline
    & & &$2W$ & & & &$2L$ & & & &$2W$ & & & \\
    \hline
    & &$2W$ & & & &$2L$ & &$2L$ & & & &$2W$ & & \\
    \hline
    &$2W$ & & & &$3L$ & & & &$3L$ & & & &$2W$ & \\
    \hline
    $3W$ & & &$2L$ & & & &$3W$ & & & &$2L$ & & &$3W$ \\
    \hline
    \end{tabular}
    \caption{Standard \textsc{Scrabble} board}
    \label{table:scrabble_board}
\end{figure}
There are 2 general sets of acceptable words, named OCTWL and SOWPODS. These 2 sets were developed specially for \textsc{Scrabble} so that there are only words of 2-15 characters. OCTWL is generally used in the USA, Canada, and Thailand while other countries are using SOWPODS. There are differences in the number of words, as shown in Table~\ref{table:scrabble_acceptable_words_distribution}.
\begin{table}[ht]
    \caption{Acceptable words distribution in \textsc{Scrabble}}
    \label{table:scrabble_acceptable_words_distribution}
    \centering
    \begin{tabular}{p{2.2cm} p{4.0cm} l} 
    \Xhline{4\arrayrulewidth}
    Set &OCTWL &SOWPODS \\ 
    \hline
    Usage &USA, Canada, Thailand &Others \\
    Total Word &187632 &267751 \\
    \Xhline{4\arrayrulewidth}
    \end{tabular}
\end{table}

Table~\ref{table:scrabble_population} shows the population distribution of players from \textsc{cross-tables} \cite{url:cross_tables}, the unofficial online \textsc{Scrabble} resource. Obviously, there is a large difference between those who are the native speaker and those who are not. Then, we hypothesize that current setting of \textsc{Scrabble} is more attractive for players that have sufficient English knowledge, than most language learners.
\begin{table}[ht]
    \caption{Population distribution of \textsc{Scrabble} players in \textsc{cross-tables}}
    \label{table:scrabble_population}
    \centering
    \begin{tabular}{p{1.8cm} p{2.6cm} r r} 
    \Xhline{4\arrayrulewidth}
    Country &Official language &Players &\%~~ \\ 
    \hline
    Barbados &English, Bajan    &2~~    &0.149 \\
    Canada &English, French     &293~~  &21.8331 \\
    Israel &Hebrew, Arabic      &1~~    &0.0745 \\
    Thailand &Thai              &3~~    &0.2235 \\
    USA &English                &1041~~ &77.5708 \\
    Unknown &Unknown            &2~~    &0.149 \\
    \Xhline{4\arrayrulewidth}
    \end{tabular}
\end{table}
\section{Game Refinement Measure}
\label{section:game_refinement_measure}
This section gives a short description of game refinement theory. A general model of game refinement was proposed based on the concept of game progress and game information progress \cite{paper:mathematical_theory_of_game_refinement}. It bridges a gap between boardgames and sports games.

\subsection{Game Progress Model}
The 'game progress' is twofold. One is game speed or scoring rate, while another one is game information progress which focuses on the game outcome. Game information progress presents the degree of certainty of the game's results in time or in steps. Having full information of the game progress, i.e. after its conclusion, game progress $x(t)$ will be given as a linear function of time $t$ with $0 \leq t \leq t_k$ and $0 \leq x(t) \leq x(t_k)$, as shown in Eq.~(\ref{equation:game_refinement_history_1}).
\begin{equation}
\label{equation:game_refinement_history_1}
	x(t) = \frac{x(t_k)}{t_k} ~ t
\end{equation}
However, the game information progress given by Eq.~(\ref{equation:game_refinement_history_1}) is unknown during the in-game period. The presence of uncertainty during the game, often until the final moments of a game, reasonably renders game progress exponential. Hence, a realistic model of game information progress is given by Eq.~(\ref{equation:game_refinement_history_2}).
\begin{equation}
\label{equation:game_refinement_history_2}
    x(t) =  x(t_k) (\frac{t}{t_k})^n
\end{equation}
Here $n$ stands for a constant parameter which is given based on the perspective of an observer of the game that is considered. Then the acceleration of the game information progress is obtained by deriving Eq.~(\ref{equation:game_refinement_history_2}) twice. Solving it at $t = t_k$, we have Eq.~(\ref{equation:game_refinement_history_3}).
\begin{equation}
\label{equation:game_refinement_history_3}
	x''(t_k) = \frac{x(t_k)}{(t_k)^n} t^{n-2} ~ n(n-1) = \frac{x(t_k)}{(t_k)^2} ~ n(n-1)
\end{equation}
It is assumed in the current model that game information progress in any type of game is encoded and transported in our brains. We do not yet know about the physics of information in the brain, but it is likely that the acceleration of information progress is subject to the forces and laws of physics. Therefore, we expect that the larger the value $\frac{x(t_k)}{(t_k)^2}$, the more exciting the game becomes, due in part to the uncertainty of the game outcome. Thus, we use its root square, $\frac{\sqrt{x(t_k)}}{t_k}$, as a game refinement measure for the game under consideration. We call it $GR$ value for short, also call $x(t_k)$ and $t_k$ as $G$ and $T$ respectively, as shown in Eq.~(\ref{equation:game_refinement_history_4}).
\begin{equation}
\label{equation:game_refinement_history_4}
	GR = \frac{\sqrt{G}}{T}
\end{equation}
In the previous works, the game progress model has been applied to various sports games \cite{paper:game_refinement_theory_score_limit_games} to verify its effectiveness. The appropriate zone of game refinement measure range from 0.07 to 0.08. The game progress model has been expanded to other domains such as multiplayer card games \cite{paper:game_refinement_theory_and_multiplayer_game_uno} and video games \cite{paper:quantifying_engagement_of_various_games}. 
We show, in Table~\ref{table:game_refinement_comparison_game_progress_model}, the results of measures of game refinement for some games.
\begin{table}[ht]
    \caption{Comparison of game refinement values for some games}
    \label{table:game_refinement_comparison_game_progress_model}
    \centering
    \begin{tabular}{l c c c}
    \Xhline{4\arrayrulewidth}
        &Successful shoot (G) &Attempt (T) & GR \\ 
    \hline
    Soccer &2.64 &22 &0.073 \\
    Basketball &36.38 &82.01 &0.073 \\
    UNO &0.976 &12.684 &0.078 \\
    Badminton &46.336 &79.344 &0.086 \\
    Table Tennis &54.863 &96.465 &0.077 \\
    DotA &68.6 &106.2 &0.078 \\
    \Xhline{4\arrayrulewidth}
    \end{tabular}
\end{table}
\subsection{Swing Model}
In scoring boardgames like \textsc{Scrabble}, swing, a state transition of advantage during the game progress is considered as successful shoot, and game length as attempt respectively. Let $S$ and $N$ be the average number of swings and the game length, respectively. Then the refinement measure in the swing model is given by Eq.~(\ref{equation:game_refinement_swing}).
\begin{equation}
    GR = \frac{\sqrt{S}}{N}
\label{equation:game_refinement_swing}
\end{equation}
\section{Another Measure from Educational Perspective}
\label{section:education_dimension}
This section gives a new measure from the educational perspective given by focusing on a balance between complexity and learning efficiency.
\subsection{Complexity}
The measure of search-space complexity or complexity \cite{Allis1994} indicates the total possible in the game represented on the natural logarithm scale. Let $B$ and $D$ be the average branching factor and average game length respectively. The complexity is obtained by Eq.~(\ref{equation:complexity}).
\begin{equation}
    C = D\ln{B}
\label{equation:complexity}
\end{equation}
Complexity measure can express the complexity of the game from the viewpoint of players. The player who has the ability to handle problems with higher complexity would think wider and deeper, thus have a better solution and understanding the nature of the game. The complexity of some existing games from the viewpoint of experts are shown in Table~\ref{table:complexity_comparison}. According to the history of game artificial intelligence development, chess computer Deep Blue won a world champion Garry Kasparov in May 1997 \cite{paper:deep_blue_artificial_intelligence}. Then, Go computer AlphaGo won a world champion Ke Jie in May 2017 \cite{paper:mastering_the_game_of_go}. The difficulty in artificial intelligence development obviously shows that complexity of Go is much more than that of chess. Similarly, the complexity of chess is much more than that of Tic-tac-toe.
\begin{table}[ht]
    \caption{Comparison of complexity for some board games}
    \label{table:complexity_comparison}
    \centering
    \begin{tabular}{p{2cm} c c c}
    \Xhline{4\arrayrulewidth}
                & Branching factor (B)  & Game length (D)   & Complexity (C) \\ 
    \hline
    Tic-tac-toe & $\leq$9               & $\leq$9           & $\leq$19.775 \\
    Chess       & 35                    & 80                & 284.428 \\
    Go          & 250	                & 208               & 1148.464 \\
    \Xhline{4\arrayrulewidth}
    \end{tabular}
\end{table}
\subsection{Learning Coefficient}
From the experiments performed with different size of dictionary and different player's model, we found that the complexity measure has the linear relation with AI knowledge base. It enables to calculate the slope per each dictionary size. Let $d$ and $t$ be the custom dictionary size and total words in dictionary respectively. The total words in custom dictionary $d'$ can be obtained by Eq.~(\ref{equation:complexity_slope_proof_1}).
\begin{equation}
\label{equation:complexity_slope_proof_1}
	d' = td
\end{equation}
Let $p$ and $x$ be the current player knowledge base and the newly learned words respectively. The new player knowledge base $p'$ can be obtained by Eq.~(\ref{equation:complexity_slope_proof_2}).
\begin{equation}
\label{equation:complexity_slope_proof_2}
	p' = p + \frac{x}{d'} = p + \frac{x}{td}
\end{equation}
By the definition of slope $m$, it can be obtained by the difference of complexity in proportional to the difference of player knowledge base, as shown by Eq.~(\ref{equation:complexity_slope_proof_3}).
\begin{equation}
\label{equation:complexity_slope_proof_3}
    m = \frac{\Delta c}{\Delta p} = \frac{\Delta c}{p' - p} = \frac{\Delta c}{p + \frac{x}{td} - p} = \frac{\Delta c}{\frac{x}{td}} 
\end{equation}
Thus we have learning coefficient (say $\Delta c$), as shown by 
Eq.~(\ref{equation:complexity_slope_proof_4}).
\begin{equation}
\label{equation:complexity_slope_proof_4}
    \Delta c = \frac{mx}{td}
\end{equation}
The newly learned words $x$ and total words $d$ are constant. To maximize the benefit, we need to maximize the increment of $\Delta c$ since it represents the improvement of a learner with the same amount of newly learned words. In short, we need to maximize $\frac{m}{d}$, which we define in this study as learning coefficient.

\section{Assessment and Discussion}
\label{section:assessment_and_discussion}
%
%
Experiments are conducted by simulating the \textsc{Scrabble} matches between AI with the various knowledge base, using various dictionary sizes. 

\subsection{Possible Enhancement with focus on Complexity}

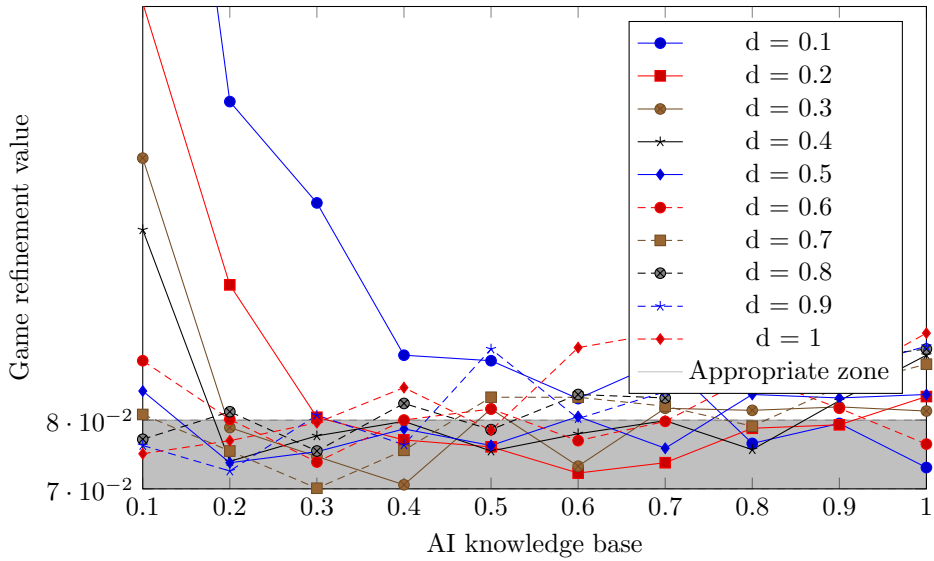
\begin{figure}[h!]
\begin{tikzpicture}
\begin{axis}[
    width=12cm,
    height=8cm,
    xlabel={AI knowledge base},
    ylabel={Game refinement value},
    xmin=0.1, xmax=1,
    ymin=0.07, ymax=0.14,
    xtick={0, 0.1, 0.2, 0.3, 0.4, 0.5, 0.6, 0.7, 0.8, 0.9, 1},
    ytick={0, 0.07, 0.08},
    legend pos=north east,
    ymajorgrids=true,
    grid style=dashed,
    cycle list name=color
]
\addplot
    coordinates {
    (0.1, 0.2204) (0.2, 0.1262) (0.3, 0.1115) (0.4, 0.0894) (0.5, 0.0886) (0.6, 0.0831) (0.7, 0.0887) (0.8, 0.0766) (0.9, 0.0795) (1, 0.0731) 
    };
\addlegendentry{d = 0.1}

\addplot
    coordinates {
    (0.1, 0.1405) (0.2, 0.0996) (0.3, 0.0804) (0.4, 0.0771) (0.5, 0.0761) (0.6, 0.0723) (0.7, 0.0738) (0.8, 0.0788) (0.9, 0.0793) (1, 0.0834) 
    };
\addlegendentry{d = 0.2}

\addplot
    coordinates {
    (0.1, 0.118) (0.2, 0.0789) (0.3, 0.0747) (0.4, 0.0706) (0.5, 0.0816) (0.6, 0.0733) (0.7, 0.0817) (0.8, 0.0814) (0.9, 0.0819) (1, 0.0813) 
    };
\addlegendentry{d = 0.3}

\addplot
    coordinates {
    (0.1, 0.1076) (0.2, 0.074) (0.3, 0.0777) (0.4, 0.0798) (0.5, 0.0755) (0.6, 0.078) (0.7, 0.0799) (0.8, 0.0757) (0.9, 0.0828) (1, 0.0894) 
    };
\addlegendentry{d = 0.4}

\addplot
    coordinates {
    (0.1, 0.0842) (0.2, 0.0738) (0.3, 0.0754) (0.4, 0.0787) (0.5, 0.0763) (0.6, 0.0805) (0.7, 0.0759) (0.8, 0.0837) (0.9, 0.0832) (1, 0.0837) 
    };
\addlegendentry{d = 0.5}

\addplot
    coordinates {
    (0.1, 0.0886) (0.2, 0.0801) (0.3, 0.0739) (0.4, 0.08) (0.5, 0.0816) (0.6, 0.077) (0.7, 0.0798) (0.8, 0.0864) (0.9, 0.0817) (1, 0.0765) 
    };
\addlegendentry{d = 0.6}

\addplot
    coordinates {
    (0.1, 0.0808) (0.2, 0.0755) (0.3, 0.0701) (0.4, 0.0756) (0.5, 0.0833) (0.6, 0.0833) (0.7, 0.082) (0.8, 0.0791) (0.9, 0.0848) (1, 0.0881) 
    };
\addlegendentry{d = 0.7}

\addplot
    coordinates {
    (0.1, 0.0772) (0.2, 0.0812) (0.3, 0.0755) (0.4, 0.0824) (0.5, 0.0786) (0.6, 0.0837) (0.7, 0.0832) (0.8, 0.0892) (0.9, 0.0887) (1, 0.0903) 
    };
\addlegendentry{d = 0.8}

\addplot
    coordinates {
    (0.1, 0.0763) (0.2, 0.0726) (0.3, 0.0807) (0.4, 0.0764) (0.5, 0.0903) (0.6, 0.0802) (0.7, 0.0841) (0.8, 0.0911) (0.9, 0.0872) (1, 0.0907) 
    };
\addlegendentry{d = 0.9}

\addplot
    coordinates {
    (0.1, 0.0751) (0.2, 0.077) (0.3, 0.0796) (0.4, 0.0847) (0.5, 0.0785) (0.6, 0.0905) (0.7, 0.0927) (0.8, 0.0875) (0.9, 0.0856) (1, 0.0926) 
    };
\addlegendentry{d = 1}

\addplot[
    color=black,
    opacity=0.25,
    name path=A
    ] 
    plot coordinates {
    (0, 0.07)(1, 0.07)
    };
\addlegendentry{Appropriate zone}

\addplot[
    color=black,
    opacity=0.25,
    name path=B
    ]
    plot coordinates {
    (0,0.08)(1,0.08)
    };
\addplot[
    color=black,
    fill opacity=0.25
    ] 
    fill between[of=A and B];
\end{axis}
\end{tikzpicture}
\caption{Measures of game refinement and AI knowledge base: standard \textsc{Scrabble} matches between AIs with different knowledge base on various dictionary size}
\label{figure:15x15_game_refinement}
\end{figure}

\begin{figure}[h!]
\begin{tikzpicture}
\begin{axis}[
    width=12cm,
    height=8cm,
    xlabel={AI knowledge base},
    ylabel={Game refinement value},
    xmin=0.1, xmax=1,
    ymin=0.07, ymax=0.14,
    xtick={0, 0.1, 0.2, 0.3, 0.4, 0.5, 0.6, 0.7, 0.8, 0.9, 1},
    ytick={0, 0.07, 0.08},
    legend pos=north east,
    ymajorgrids=true,
    grid style=dashed,
    cycle list name=color
]
\addplot
    coordinates {
    (0.1, 0.2352) (0.2, 0.1641) (0.3, 0.1185) (0.4, 0.1255) (0.5, 0.0997) (0.6, 0.0864) (0.7, 0.0971) (0.8, 0.0918) (0.9, 0.0978) (1, 0.0933) 
    };
\addlegendentry{d = 0.1}

\addplot
    coordinates {
    (0.1, 0.1371) (0.2, 0.1029) (0.3, 0.0898) (0.4, 0.0897) (0.5, 0.0784) (0.6, 0.0782) (0.7, 0.0804) (0.8, 0.0792) (0.9, 0.082) (1, 0.0818) 
    };
\addlegendentry{d = 0.2}

\addplot
    coordinates {
    (0.1, 0.1309) (0.2, 0.0989) (0.3, 0.0883) (0.4, 0.0789) (0.5, 0.0767) (0.6, 0.0845) (0.7, 0.0789) (0.8, 0.0821) (0.9, 0.0784) (1, 0.0768) 
    };
\addlegendentry{d = 0.3}

\addplot
    coordinates {
    (0.1, 0.0955) (0.2, 0.0855) (0.3, 0.0768) (0.4, 0.0679) (0.5, 0.0762) (0.6, 0.0816) (0.7, 0.0756) (0.8, 0.0789) (0.9, 0.0815) (1, 0.0738) 
    };
\addlegendentry{d = 0.4}

\addplot
    coordinates {
    (0.1, 0.0965) (0.2, 0.0796) (0.3, 0.0803) (0.4, 0.0738) (0.5, 0.0755) (0.6, 0.0743) (0.7, 0.0728) (0.8, 0.0698) (0.9, 0.0767) (1, 0.0814) 
    };
\addlegendentry{d = 0.5}

\addplot
    coordinates {
    (0.1, 0.1025) (0.2, 0.0786) (0.3, 0.0781) (0.4, 0.0751) (0.5, 0.0783) (0.6, 0.0789) (0.7, 0.0738) (0.8, 0.0767) (0.9, 0.0782) (1, 0.0725) 
    };
\addlegendentry{d = 0.6}

\addplot
    coordinates {
    (0.1, 0.0857) (0.2, 0.0707) (0.3, 0.0779) (0.4, 0.0739) (0.5, 0.0779) (0.6, 0.0717) (0.7, 0.079) (0.8, 0.0819) (0.9, 0.0747) (1, 0.0736) 
    };
\addlegendentry{d = 0.7}

\addplot
    coordinates {
    (0.1, 0.0824) (0.2, 0.0783) (0.3, 0.0705) (0.4, 0.0792) (0.5, 0.0792) (0.6, 0.0735) (0.7, 0.0802) (0.8, 0.077) (0.9, 0.0734) (1, 0.0778) 
    };
\addlegendentry{d = 0.8}

\addplot
    coordinates {
    (0.1, 0.087) (0.2, 0.072) (0.3, 0.0767) (0.4, 0.075) (0.5, 0.0747) (0.6, 0.0777) (0.7, 0.0764) (0.8, 0.0819) (0.9, 0.078) (1, 0.0809) 
    };
\addlegendentry{d = 0.9}

\addplot
    coordinates {
    (0.1, 0.0771) (0.2, 0.0761) (0.3, 0.0787) (0.4, 0.073) (0.5, 0.0723) (0.6, 0.0788) (0.7, 0.0832) (0.8, 0.0791) (0.9, 0.0795) (1, 0.0808) 
    };
\addlegendentry{d = 1}

\addplot[
    color=black,
    opacity=0.25,
    name path=A
    ] 
    plot coordinates {
    (0.1, 0.07) (1, 0.07)
    };
\addlegendentry{Appropriate zone}

\addplot[
    color=black,
    opacity=0.25,
    name path=B
    ]
    plot coordinates {
    (0.1, 0.08) (1, 0.08)
    };
\addplot[
    color=black,
    fill opacity=0.25
    ] 
    fill between[of=A and B];
\end{axis}
\end{tikzpicture}
\caption{Measures of game refinement and AI knowledge base: 13x13 variation of \textsc{Scrabble} matches between AIs with different knowledge base on various dictionary size}
\label{figure:13x13_game_refinement}
\end{figure}
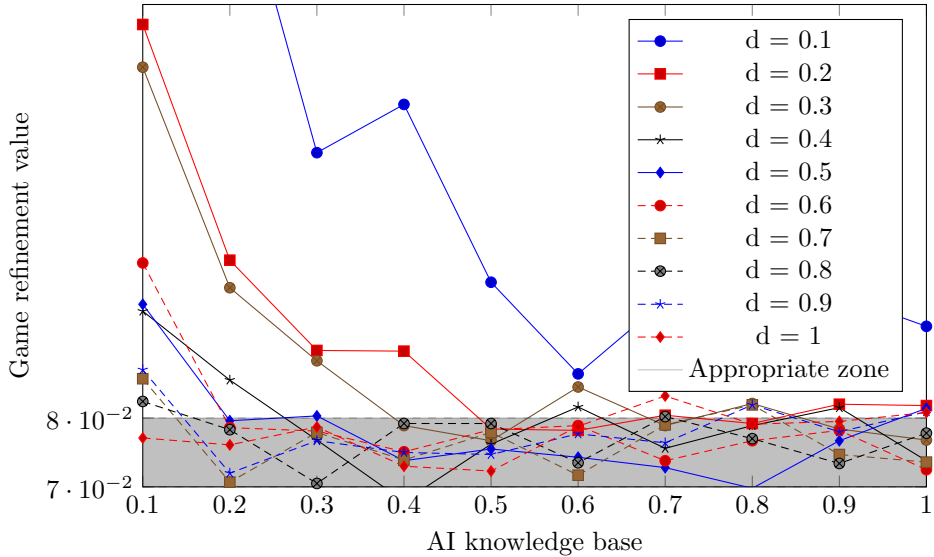

The experimental results are analyzed using game refinement measure, as shown in Fig.~\ref{figure:15x15_game_refinement}.
%
Game refinement measure of the original \textsc{Scrabble} is slightly higher than appropriate zone. This reveals that the original setting of \textsc{Scrabble} yields excess branching factors. One possible enhancement is to reduce the board size, and it is found that 13x13 board size gives the best setting. The complete board is shown in Fig.~\ref{table:scrabble_board_13x13}. This results in 24.89\% smaller compared to the standard \textsc{Scrabble}, thus can significantly reduce the branching factor. The results are shown in Fig.~\ref{figure:13x13_game_refinement}, which is much closer to the appropriate zone.
\begin{figure}[h!]
    \centering
    \begin{tabular}{|c|c|c|c|c|c|c|c|c|c|c|c|c|c|c|} 
    \hline
    $2W$ & & & &$3L$ & & & &$3L$ & & & &$2W$\\
    \hline
    &$2W$ & & & &$2L$ & &$2L$ & & & &$2W$ &\\
    \hline
    & &$2W$ & & & &$2L$ & & & &$2W$ & &\\
    \hline
    & & &$2W$ & & & & & &$2W$ & & &\\
    \hline
    $3L$ & & & &$3L$ & & & &$3L$ & & & &$3L$\\
    \hline
    &$2L$ & & & &$2L$ & &$2L$ & & & &$2L$ &\\
    \hline
    & &$2L$ & & & &$2W$ & & & &$2L$ & &\\
    \hline
    &$2L$ & & & &$2L$ & &$2L$ & & & &$2L$ &\\
    \hline
    $3L$ & & & &$3L$ & & & &$3L$ & & & &$3L$\\
    \hline
    & & &$2W$ & & & & & &$2W$ & & &\\
    \hline
    & &$2W$ & & & &$2L$ & & & &$2W$ & &\\
    \hline
    &$2W$ & & & &$2L$ & &$2L$ & & & &$2W$ &\\
    \hline
    $2W$ & & & &$3L$ & & & &$3L$ & & & &$2W$\\
    \hline
    \end{tabular}
    \caption{13x13 variant of \textsc{Scrabble} board}
    \label{table:scrabble_board_13x13}
\end{figure}
%

\subsection{Tendency of Game Refinement Measure Changes}
The interesting part is that \textsc{Scrabble} with various size of a dictionary has different game refinement tendency, as shown in Table~\ref{table:15x15_game_refinement_tendency}. Note that `Dec' and `Inc' stand for decreasing and increasing, respectively.
\begin{table}[ht]
    \caption{Tendency of game refinement measure changes for different size of dictionary}
    \label{table:15x15_game_refinement_tendency}
    \centering
    \begin{tabular}{p{2.7cm} l}
    \Xhline{4\arrayrulewidth}
    Dictionary size &GR tendency \\
    \hline
    0.01 - 0.1  & Dec \\
    0.2 - 0.6   & Dec then Inc \\
    0.7 - 0.9   & Inc \\
    1           & Inc then Dec \\
    \Xhline{4\arrayrulewidth}
    \end{tabular}
\end{table}

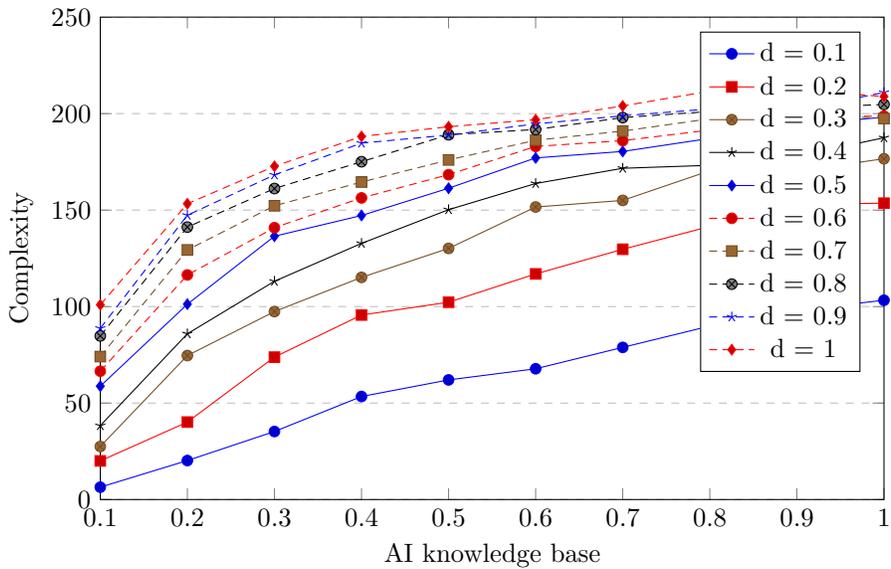
\begin{figure}[htb]
\begin{tikzpicture}
\begin{axis}[
    width=12cm,
    height=8cm,
    xlabel={AI knowledge base},
    ylabel={Complexity},
    xmin=0.1, xmax=1,
    ymin=0, ymax=250,
    xtick={0, 0.1, 0.2, 0.3, 0.4, 0.5, 0.6, 0.7, 0.8, 0.9, 1},
    ytick={0, 50, 100, 150, 200, 250},
    legend pos=north east,
    ymajorgrids=true,
    grid style=dashed,
    cycle list name=color
]
\addplot
    coordinates {
    (0.1, 6.4198) (0.2, 20.2083) (0.3, 35.2646) (0.4, 53.4054) (0.5, 62.0025) (0.6, 67.7768) (0.7, 78.855) (0.8, 89.8998) (0.9, 98.0648) (1, 103.2699) 
    };
\addlegendentry{d = 0.1}

\addplot
    coordinates {
    (0.1, 20.0262) (0.2, 40.1534) (0.3, 73.8202) (0.4, 95.6239) (0.5, 102.2435) (0.6, 116.9369) (0.7, 129.7059) (0.8, 141.53) (0.9, 152.9471) (1, 153.59) 
    };
\addlegendentry{d = 0.2}

\addplot
    coordinates {
    (0.1, 27.4848) (0.2, 74.6213) (0.3, 97.4117) (0.4, 115.1401) (0.5, 130.1514) (0.6, 151.604) (0.7, 155.0226) (0.8, 170.4017) (0.9, 170.1304) (1, 176.6164) 
    };
\addlegendentry{d = 0.3}

\addplot
    coordinates {
    (0.1, 38.3081) (0.2, 85.8965) (0.3, 113.1143) (0.4, 132.7561) (0.5, 150.2232) (0.6, 163.785) (0.7, 171.668) (0.8, 173.2142) (0.9, 177.9248) (1, 187.4923) 
    };
\addlegendentry{d = 0.4}

\addplot
    coordinates {
    (0.1, 58.7438) (0.2, 101.2365) (0.3, 136.4542) (0.4, 147.209) (0.5, 161.3325) (0.6, 177.0388) (0.7, 180.3928) (0.8, 187.1519) (0.9, 194.0312) (1, 198.1639) 
    };
\addlegendentry{d = 0.5}

\addplot
    coordinates {
    (0.1, 66.5162) (0.2, 116.3982) (0.3, 140.903) (0.4, 156.3169) (0.5, 168.3566) (0.6, 182.9952) (0.7, 185.9932) (0.8, 191.4725) (0.9, 198.5544) (1, 198.6712) 
    };
\addlegendentry{d = 0.6}

\addplot
    coordinates {
    (0.1, 74.0912) (0.2, 129.3638) (0.3, 152.2268) (0.4, 164.5299) (0.5, 175.9804) (0.6, 186.2004) (0.7, 190.9443) (0.8, 197.3241) (0.9, 200.2789) (1, 197.414) 
    };
\addlegendentry{d = 0.7}

\addplot
    coordinates {
    (0.1, 84.7892) (0.2, 141.1889) (0.3, 161.1653) (0.4, 175.0712) (0.5, 189.0449) (0.6, 191.7569) (0.7, 197.9346) (0.8, 201.1047) (0.9, 202.8148) (1, 204.7567) 
    };
\addlegendentry{d = 0.8}

\addplot
    coordinates {
    (0.1, 88.7559) (0.2, 147.3221) (0.3, 168.2361) (0.4, 184.7118) (0.5, 188.9076) (0.6, 194.681) (0.7, 198.8788) (0.8, 202.4208) (0.9, 202.0299) (1, 210.9456) 
    };
\addlegendentry{d = 0.9}

\addplot
    coordinates {
    (0.1, 100.927) (0.2, 153.2947) (0.3, 172.6909) (0.4, 188.158) (0.5, 193.2091) (0.6, 196.8328) (0.7, 203.9766) (0.8, 211.658) (0.9, 212.1199) (1, 208.8595) 
    };
\addlegendentry{d = 1}

\end{axis}
\end{tikzpicture}
\caption{Complexity and AI knowledge base: standard \textsc{Scrabble} matches between AIs with different knowledge base on various dictionary size}
\label{figure:15x15_complexity}
\end{figure}

From the earlier study \cite{paper:mathematical_theory_of_game_refinement}, a measure of game refinement would reflect the balance between chance and skill in playing the game under consideration. Higher game refinement value means that chance is a stronger factor than skill. Game refinement tendency would indicate the user experience of the game. If it is increasing, we know that chance has more effect for a match-up between expert players, and less for novice players respectively. Novice used to enjoy this type of game as a skill-based game, but once they become an expert player, only individual skill is not enough to beat the opponent since there are some other unexpected factors towards a game, e.g., chance, teamwork and imperfect information. This would offer the fun-game experience. 
On the contrary, tendency of decreasing would indicate the competitive-game experience. In case where tendency is  decreasing-then-increasing, it combines both experiences in the different phase. At the beginning, players may feel the competitive-game experience, which is followed by the fun-game experience. 
On the other hand, the results using complexity measure are shown in Fig.~\ref{figure:15x15_complexity}.
%

\subsection{Learning Coefficient and Summary}

\begin{figure}[h!]
\begin{tikzpicture}
\begin{axis}[
    width=12cm,
    height=8cm,
    xlabel={Dictionary size},
    ylabel={Complexity slope per dictionary size},
    xmin=0, xmax=1,
    ymin=50, ymax=1350,
    xtick={0, 0.1, 0.2, 0.3, 0.4, 0.5, 0.6, 0.7, 0.8, 0.9, 1},
    ytick={200, 400, 600, 800, 1000, 1200},
    legend pos=north east,
    ymajorgrids=true,
    grid style=dashed,
    cycle list name=color
]
\addplot
    coordinates {
    (0.01, 854.98)	(0.02, 803.97)	(0.03, 997.4433)	(0.04, 1333.335)	(0.05, 1231.206)	(0.06, 1280.6317)	(0.07, 930.6857)	(0.08, 1122.5488)	(0.09, 1051.3033)	(0.1, 1073.907)	(0.2, 741.5515)	(0.3, 508.4443)	(0.4, 366.311)	(0.5, 275.526)	(0.6, 214.2428)	(0.7, 166.3429)	(0.8, 135.0066)	(0.9, 114.6033)	(1, 98.7323)
    };
\end{axis}
\end{tikzpicture}
\caption{Relation between dictionary size and corresponding learning coefficient}
\label{figure:15x15_complexity_slope}
\end{figure}
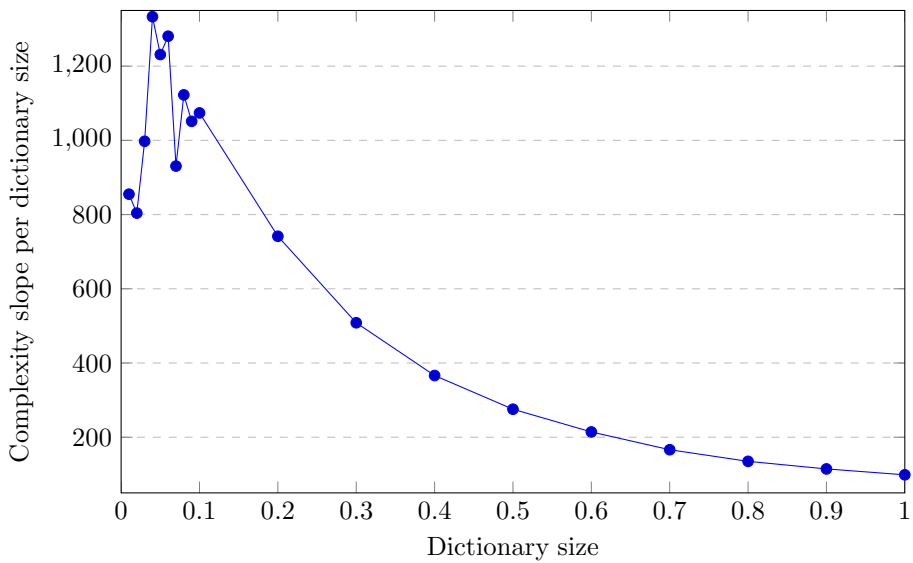

For every dictionary size, we have compared the learning coefficient as shown in Fig.~\ref{figure:15x15_complexity_slope}. 
%
The peak of the highest complexity slope per dictionary size is at 4\% of dictionary size. Despite that, the corresponding game refinement measure is relatively uncomfortable. The good balance between entertainment and education would be around 10\% to 20\% of dictionary size.
%
We show, in Table~\ref{table:conclusion}, the summary of our results in this study. 
\begin{table}[h!]
    \caption{Summary of analyzing \textsc{Scrabble} using game refinement measure and learning coefficient}
    \label{table:conclusion}
    \centering
    \begin{tabular}{l c c c l r}
    \Xhline{4\arrayrulewidth}
    Variation   & Board size & d    & GR                & GR tendency & Learning coefficient \\
    \hline
    Standard        & 15x15 & 1     & 0.0751 -- 0.0926   &Inc then Dec &98.7323 \\
    Entertainment   & 13x13 & 1     & 0.0771 -- 0.0808   &Dec then inc &140.9987 \\
    Education       & 15x15 & 0.04  & 0.0951 -- 0.3843   &Dec &1333.335 \\
    Balance         & 15x15 & 0.1   & 0.0731 -- 0.2204   &Dec &1073.907 \\
    \Xhline{4\arrayrulewidth}
    \end{tabular}
\end{table}

\section{Concluding Remarks}
\label{section:conclusion}
According to the study, \textsc{Scrabble} players would feel fun-game experience (i.e., entertainment) more than the educational experience. We proposed three ways of possible improvement: entertainment enhancement, education enhancement, and the good balance between them.
\textsc{Scrabble} yields excessive branching factors. This results in that game refinement measures are higher than the appropriate zone. Entertainment enhancement can be done by reducing the standard board size (15x15) to 13x13. This can improve game refinement values significantly, specifically for native speakers to enjoy the competitive environment.
On the other hand, we can also enhance \textsc{Scrabble} with the educational dimension. For this purpose we proposed two new models with focus on complexity and learning coefficient. Educational enhancement can be done by maximizing the learning coefficient value, while the good balance between entertainment and enhancement can be found by trading off.

\subsubsection*{Acknowledgments.}
This research is funded by a grant from the Japan Society for the Promotion of Science, in the framework of the Grant-in-Aid for Challenging Exploratory Research.

\end{document}